\documentclass[10pt,twocolumn,letterpaper]{article}

\usepackage{wacv}              

\usepackage{multirow}
\usepackage{colortbl}
\usepackage{float}
\usepackage{todonotes}
\usepackage{pgfplots}
\usepackage{booktabs}
\usepackage{adjustbox}
\usepackage{tikz}
\usepackage{graphicx}
\usepackage{xcolor}
\usepackage{amsmath, amssymb}
\usepackage[ruled]{algorithm2e}
\usepackage{algorithmicx}
\definecolor{darkgreen}{RGB}{0,150,0}
\usetikzlibrary{plotmarks}

\pgfplotsset{compat=1.18} 

\makeatletter
\def\thickhline{%
\noalign{\ifnum0=`}\fi\hrule \@height \thickarrayrulewidth \futurelet
\reserved@a\@xthickhline}
\def\@xthickhline{\ifx\reserved@a\thickhline
            \vskip\doublerulesep
            \vskip-\thickarrayrulewidth
            \fi
    \ifnum0=`{\fi}}
\makeatother

\usepackage{xcolor}
\usepackage{pifont}
\newcommand{\xmark}{\ding{55}}
\definecolor{mygreen}{RGB}{0,128,0}

\usepackage{amsmath, amssymb}
\usepackage[ruled]{algorithm2e}
\usepackage{algorithmicx}

\usepackage{subcaption}
\usepackage{wrapfig}

\definecolor{wacvblue}{rgb}{0.21,0.49,0.74}

\usepackage[pagebackref,breaklinks,colorlinks,allcolors=wacvblue]{hyperref}

\def\wacvPaperID{1935} 
\def\confName{WACV}
\def\confYear{2026}

\title{Towards Streaming LiDAR Object Detection with Point Clouds as Egocentric Sequences}

\author{Mellon M. Zhang \quad Glen Chou\textsuperscript{\dag} \quad Saibal Mukhopadhyay\textsuperscript{\dag} \\
Georgia Institute of Technology\\
{\tt\small \{meilongz,chou\}@gatech.edu, saibal@ece.gatech.edu}
}

\begin{document}
\maketitle
\def\thefootnote{\dag}\footnotetext{Equal advising.}
\begin{abstract}

Accurate and low-latency 3D object detection is essential for autonomous driving, where safety hinges on both rapid response and reliable perception. While rotating LiDAR sensors are widely adopted for their robustness and fidelity, current detectors face a trade-off: streaming methods process partial polar sectors on the fly for fast updates but suffer from limited visibility, cross-sector dependencies, and distortions from retrofitted Cartesian designs, whereas full-scan methods achieve higher accuracy but are bottlenecked by the inherent latency of a LiDAR revolution. We propose \textbf{Polar-Fast-Cartesian-Full (PFCF)}, a hybrid detector that combines fast polar processing for intra-sector feature extraction with accurate Cartesian reasoning for full-scene understanding. Central to PFCF is a custom Mamba SSM-based streaming backbone with dimensionally-decomposed convolutions that avoids distortion-heavy planes, enabling parameter-efficient, translation-invariant, and distortion-robust polar representation learning. Local sector features are extracted via this backbone, then accumulated into a sector feature buffer to enable efficient inter-sector communication through a full-scan backbone. PFCF establishes a new Pareto frontier on the Waymo Open dataset, surpassing prior streaming baselines by 10\% mAP and matching full-scan accuracy at twice the update rate. Code is available at \href{https://github.com/meilongzhang/Polar-Hierarchical-Mamba}{https://github.com/meilongzhang/Polar-Hierarchical-Mamba}.
\end{abstract}
\vspace{-10pt}  
\vspace{-5pt}
\section{Introduction}
\label{sec:intro}
\begin{figure}[ht]
\vspace{-4mm}
  \centering
  \includegraphics[width=0.7\linewidth]{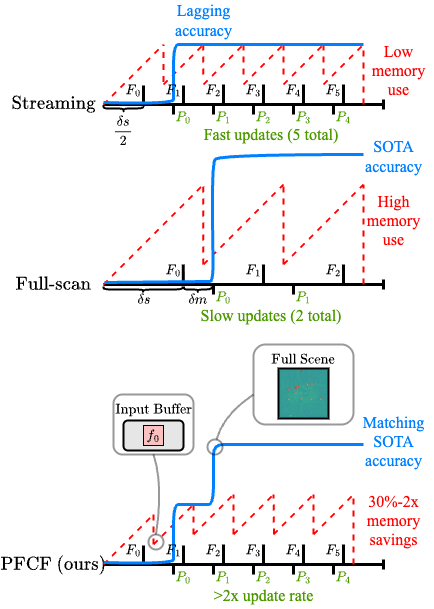}
  \vspace{-3mm}
  \caption{\textbf{Comparison of Perception Paradigms.} With the introduction of a sector feature buffer, PFCF combines the rapid updates and low peak memory utilization of streaming methods with the accuracy of full-scan methods. The solid blue lines correspond to accuracy, the dashed red lines correspond to memory usage. The $F$ values are inputs, the $P$ values are predictions, $\delta s$ is the sensor latency and $\delta m$ is the model latency.}
  \label{fig:paradigm}\vspace{-15pt}
\end{figure}

\vspace{-5pt}
Object detection is crucial for autonomous vehicles (AVs), demanding high accuracy, reactivity, and efficiency. Most AVs use LiDAR sensors, which excel in low-light and long-range scenarios. 
Scanning LiDAR sensors, which continuously rotate while outputting new partial sectors of points on the fly, are favored due to their high visual fidelity and long range. LiDAR perception methods \cite{VoxelNet, PointNet, PointPillar, SECOND, CenterPoint, FSD, PointRCNN} have typically followed a full-scan approach, aggregating multiple partial sectors together into full point clouds and processing them \cite{nuScenes, Waymo, Argoverse} in a single pass, leading to significant sensor latency -- on the order of hundreds of milliseconds -- making real-time detection challenging. Waiting for full LiDAR point clouds before processing reduces the update rate (throughput) by more than 2x (Fig. \ref{fig:paradigm}).

Recent work has explored streaming approaches that process partial LiDAR sectors as they arrive rather than waiting to aggregate a full point cloud \cite{strobe, han, polarstream, partner}. This cuts the sensor latency by anywhere from 2-8x and allows for significantly faster prediction updates, as shown in the top of Fig. ~\ref{fig:paradigm}. Initial streaming methods inherited the Cartesian coordinate system from full-scan processing, but polar coordinates have since emerged as the de-facto streaming representation due to memory efficiency \cite{polarstream}, robustness to resolution changes \cite{partner}, and fine-grained temporal awareness. While this combination yields perception systems with 2–4x the throughput of full-scan methods, a notable accuracy gap remains. This gap is created by two main issues: \textbf{(1)} processing partial sectors inevitably means a reduction in captured dependencies, as traditional mechanisms such as convolutions cannot propagate information across separate data packets, and \textbf{(2)} despite polar coordinates being superior for streaming, the architecture choices for streaming backbones have not been adapted from typical convolution-based methods to support this system. We refer to these two issues as 1) the information gap and 2) the architecture gap. 

In streaming, raw LiDAR points from small angular sectors are processed directly from UDP packets, making full-scan methods infeasible due to missing cross-sector dependencies. Streaming methods add information from auxiliary sensors \cite{strobe} or stateful memory \cite{polarstream}, but these are insufficient to close the information gap; thus, streaming methods are less accurate than full-scan despite faster processing.

Although streaming methods now favor the polar representation for efficiency, most still use convolution-based architectures \cite{SPConv}. 
However, in polar space, objects of equal size appear differently depending on their distance from the ego vehicle, breaking the translation invariance assumed by convolutions and compounding polar distortion. 
Previous works try to mitigate this issue via expensive post-processing mechanisms based on attention \cite{partner} or learnable interpolations \cite{polarstream}, but these post-hoc measures remain insufficient to close the architecture gap.

These gaps demand new architectures and representations. 
One option is the Mamba state space model \cite{mamba, mamba2}, which has strong performance across diverse streaming modalities \cite{mambaspike, asr, duplexmamba} and near-linear inference scaling without relying on translation invariance, due to its representation of data as one-dimensional sequences. However, streaming LiDAR presents a unique challenge: it combines sequential structure with complex 3D spatial geometry, making existing Mamba designs for purely sequential data ill-suited. In essence, while Mamba excels at modeling temporal sequences, its ability to capture spatial structure -- especially in 3D -- is still limited \cite{voxelmamba}. Recent efforts have applied Mamba to LiDAR perception \cite{unimamba, voxelmamba}, but only in the full-scan setting, treating Mamba as a generic processing block akin to convolution or attention. To enhance spatial awareness, these methods serialize point clouds using geometric curves like Hilbert \cite{hilbert} or Z-order \cite{zcurve} curves, and incorporate handcrafted positional encodings \cite{voxelmamba} with learnable mappings. However, these methods are fragile to resolution changes and transfer poorly to streaming because curve patterns must be recomputed and positional encodings use sliding windows assuming translation invariance. 

We propose a novel hybrid streaming detector, \textbf{Polar-Fast-Cartesian-Full (PFCF)} that addresses both the information and architecture gaps faced by previous streaming methods. To address the information gap of incomplete cross-sector spatiotemporal dependencies, we combine the strengths of streaming and full-scan methods by interfacing fast polar processing with accurate Cartesian processing via a \textit{sector feature buffer}. This enables our detector to ingest and process individual sectors independently, extracting lightweight feature representations from the polar space with a streaming backbone. Then, the sector features are stitched together within the sector feature buffer to form a complete feature map, transformed into the Cartesian space, and processed with a lightweight 2D full-scan backbone. To address the architecture gap, we propose a polar-native streaming backbone \textbf{Polar Hierarchical Mamba (PHiM)} that leverages the \textit{Mamba SSM} along with custom \textit{dimensionally-decomposed convolution operations} for a lightweight method to distortion mitigation without the need for post hoc adjustment. In this way, our detector achieves \textbf{full-scan processing at streaming speed}.

Our key contributions are as follows:

\begin{itemize}[left=0.75em]

    \item We propose a new paradigm to streaming LiDAR perception, known as Polar-Fast-Cartesian-Full (PFCF). Our PFCF detector features a hybrid architecture consisting of a fast 3D polar streaming backbone and a lightweight 2D Cartesian full-scan backbone. With the introduction of a sector feature buffer, PFCF is capable of outputting new predictions across the complete scene with every new inputted partial sector, more than doubling the update rate while maintaining high accuracy.

    \item We propose Polar Hierarchical Mamba (PHiM), the first streaming backbone designed specifically for the polar coordinate system. Unlike previous convolution-based streaming architectures which are susceptible to compounding polar distortion and require post hoc processing, PHiM is based on the Mamba state space model (SSM). PHiM features a hierarchical architecture that captures local intra-sector spatial details while also propagating inter-sector dependencies forward via the Mamba hidden state, and dimensionally-decomposed convolutions (DDCs) for spatial locality awareness while avoiding planes with high polar distortion. With these design choices, PHiM moves away from traditional translation-invariant processing.

    \item We compare our method to full-scan and streaming detectors on the Waymo Open dataset, achieving state-of-the-art streaming detection with a 10\% accuracy gain over streaming methods and over 2x the update rate of full-scan methods, setting a new Pareto frontier for efficient LiDAR detection. 
\end{itemize}
\begin{figure*}[t]
    \centering
    \includegraphics*[width=1\textwidth]{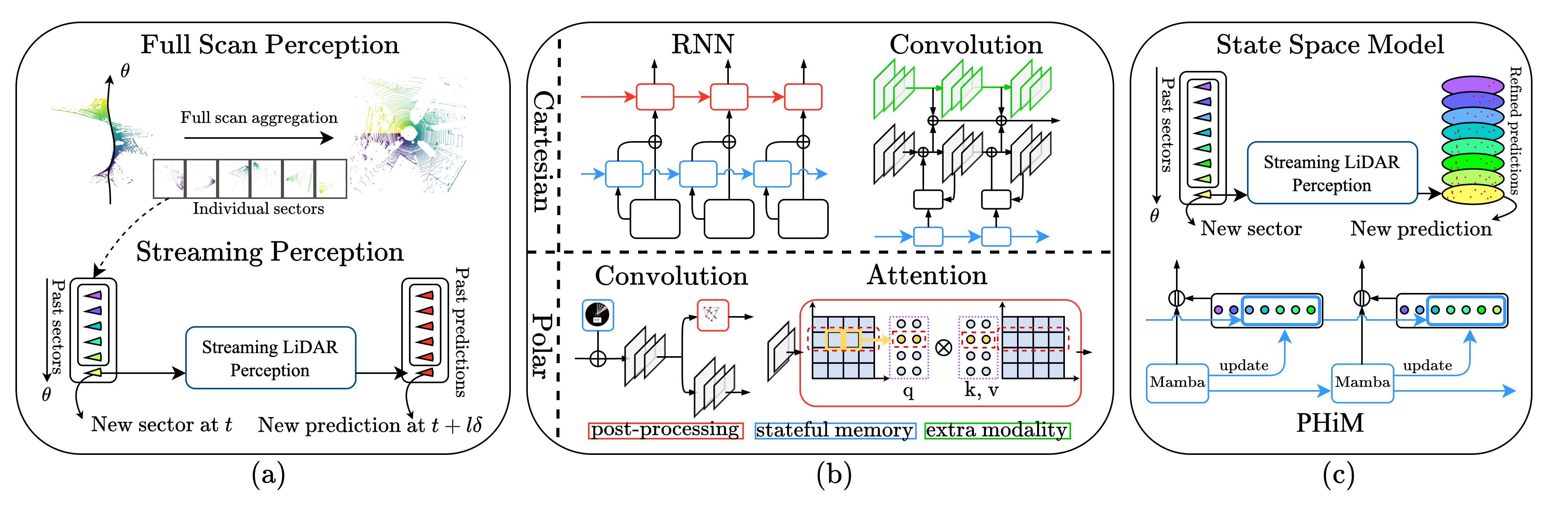}\vspace{-9pt}
    \caption{\textbf{Comparison with existing streaming works. (a)} As opposed to full scan methods that aggregate points into complete point clouds from full rotations of the LiDAR sensor, streaming methods operate on partial point cloud sectors as they are collected. Most methods make partial predictions based on these partial input sectors. \textbf{(b)} Prior streaming approaches include models operating on Cartesian coordinates using RNNs or convolutions (top), and those using polar coordinates with convolutions or attention mechanisms (bottom). These methods rely on combinations of stateful memory, post-processing, or auxiliary modalities to model spatiotemporal correlations between new and past sectors. \textbf{(c)} PHiM encodes spatiotemporal interactions directly into the hidden state of a state-space model (SSM), while storing sector-level features. This enables full-scene predictions from individual sectors and requires no additional modalities, a priori context padding, or post-processing.\vspace{-13pt}}
    \label{fig:overview}
\end{figure*}
\vspace{-8pt}
\section{Related work}
\label{sec:related}
\vspace{-5pt}

\paragraph{Full-scan methods.}

Traditional LiDAR detection aggregates partial sectors into full point cloud frames before processing, adopting translation-invariant image-processing techniques like convolutions. VoxelNet \cite{VoxelNet} voxelizes point clouds for dense 3D convolutions, while PointPillar \cite{PointPillar} and successors \cite{PillarNet, PillarNeXt} project them into BEV for lighter 2D convolutions. SECOND \cite{SECOND} advanced the field with sparse and submanifold convolutions \cite{regular_sparse, submanifold_sparse}, improving efficiency. CenterPoint \cite{CenterPoint} boosted accuracy by modeling objects as center points, inspiring refinements in localization \cite{TransFusion, CenterFormer, AFDetV2}. 

Transformer-based \cite{votr, DSVT, FastPointTransformer, FlatFormer, UVTR} models have emerged as alternatives to convolutions, designing mechanisms to mimic convolutional inductive biases. Examples include SWFormer’s local window attention \cite{SWFormer} and DSVT’s grouped attention on voxelized data \cite{DSVT}. Along a similar line, the growing popularity of Mamba in sequence modeling has also motivated its use in LiDAR perception \cite{voxelmamba, pillarmamba, unimamba}. These works serialize 3D point clouds using geometric heuristics like Hilbert \cite{hilbert} or Z-curves \cite{zcurve} and augment with positional encodings \cite{swin} to also mimic convolutional assumptions. However, these handcrafted heuristics introduce inductive bias, increase memory usage, and are poorly suited for streaming data, where the scene is observed in narrow, wedge-shaped sectors rather than full rectangular grids. 

Overall, full-scan methods achieve top accuracy on detection benchmarks \cite{Waymo, nuScenes, Argoverse, SemanticKITTI} but incur latency from collecting full point cloud frames. 
Despite speed optimizations, the waiting time for data collection remains unaddressed. Our work accelerates the update rate of full-scan methods via hybrid processing -- extracting sector features on the fly with a streaming backbone before aggregation.

\vspace{-5pt}

\paragraph{Streaming methods.}

Unlike full-scan methods that process aggregated LiDAR sweeps all at once, streaming approaches operate on partial LiDAR sectors -- angular slices of the full 360° field of view -- as they are emitted by the sensor. This enables low-latency, real-time perception, as the model can begin processing before the entire scan completes. Early methods still used Cartesian coordinates. Han \cite{han} introduces LSTMs and stateful NMS to maintain temporal context, while STROBE \cite{strobe} improves spatial memory with multi-scale aggregation and HD maps to recover missing context. 
Follow-up studies emphasize the advantages of the polar coordinate system for streaming detection, such as robustness to resolution changes \cite{polarstream, partner}, and rotation symmetry for streaming detection. However, most methods still use convolution-based designs meant for Cartesian space, causing spatial distortions that require post hoc corrections. PolarStream \cite{polarstream} uses range-stratified convolutions with bilinear sampling, and PARTNER \cite{partner} adds cross- and geometry-aware attention, yet accuracy remains limited, creating an architecture gap. 

None of the previous streaming approaches leverage state space models, nor do they challenge architectural assumptions such as translation invariance -- an assumption poorly suited for polar data. In this work, we present Polar Hierarchical Mamba (PHiM), the first Mamba-based backbone purpose-built for streaming LiDAR. PHiM models the egocentric temporal progression of LiDAR sectors directly, using dimensionally-decomposed convolutions to enhance spatial awareness while avoiding distortion-heavy planes. Our approach avoids curve-based serialization, positional embeddings, and spatial correction modules, resulting in a simpler, faster, and more general architecture. A comparison with existing streaming methods is shown in Fig.~\ref{fig:overview}, and a comparison with existing LiDAR Mamba methods is shown in Appendix \ref{app:mamba_comparison}. 
\begin{figure*}[t]
    \centering
    \includegraphics*[width=0.9\textwidth]{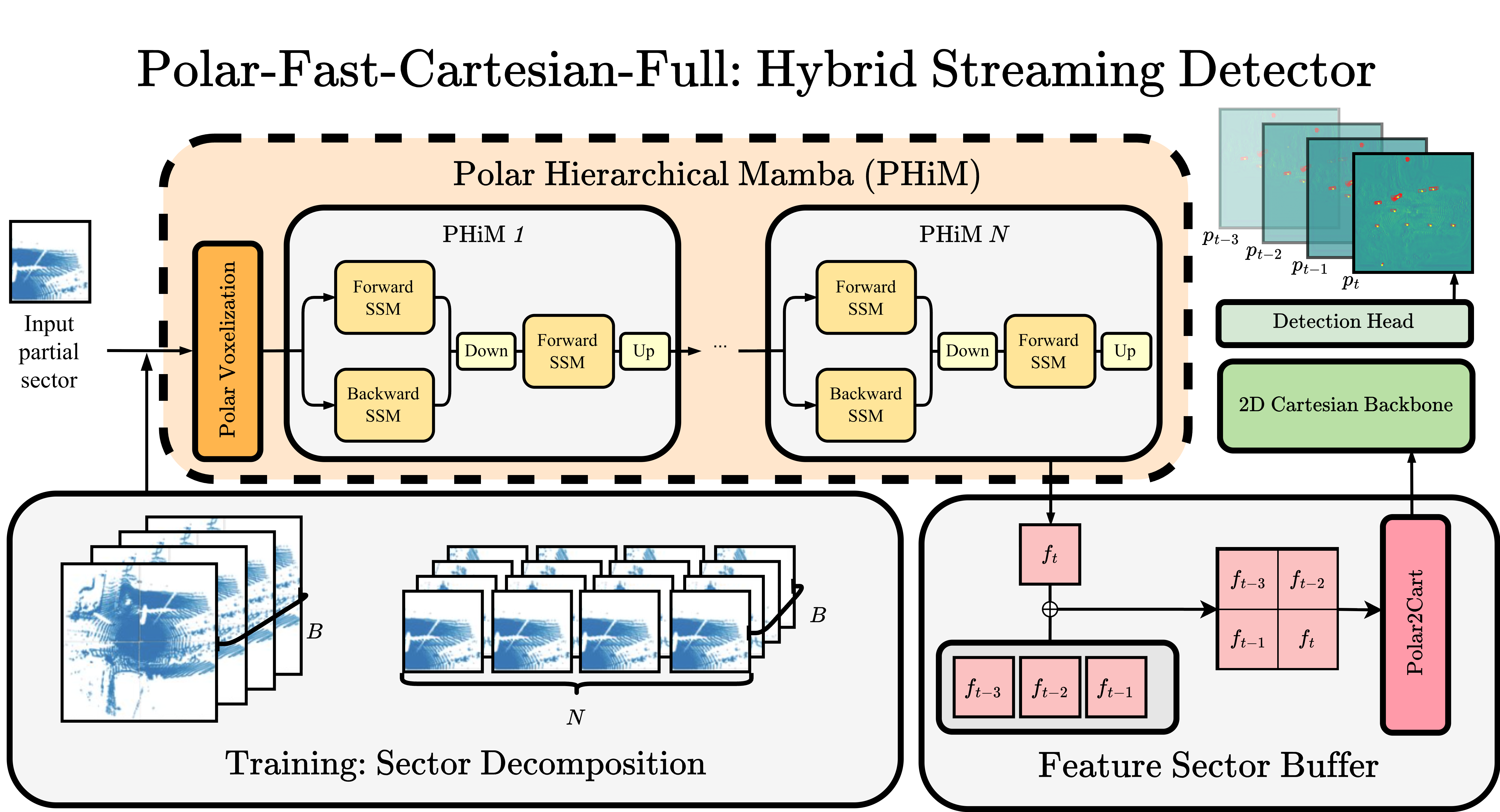}
    \caption{\textbf{Polar-Fast-Cartesian-Full (PFCF) Pipeline.} Polar-Fast-Cartesian-Full processes partial input sectors independently through a polar streaming backbone consisting of stacked Polar Hierarchical Mamba (PHiM) blocks (Fig.~\ref{fig:block_diagram}). These PHiM block saggregate local features, encode sector-level context and propagate information forward in time. Features are then stitched together with the buffered features from previous sectors, projected into the Cartesian 2D BEV space, and refined by a lightweight BEV backbone before detection via a CenterPoint head. In this way, PFCF can ingest partial sectors yet output updated predictions over the entire scene at every timestep.
    }
    \vspace{-13pt}\label{fig:pipeline}
\end{figure*}
\vspace{-8pt}
\section{Method}
\vspace{-5pt}

Our approach is motivated by a central observation: \emph{full-scan methods} tend to be highly accurate but encumbered by sensor latency, whereas \emph{streaming methods} are reactive yet often less accurate. This contrast raises a natural question: can we design a framework that combines the strengths of both paradigms to achieve the best of both worlds? Reaching this goal requires overcoming several key challenges:

\begin{enumerate}
    \item \textbf{Bidirectional refinement.} How can each partial sector contribute back to the global scene representation, allowing present observations to update and refine prior interpretations?

    \item \textbf{Temporal integration.} How can each partial sector incorporate information from previous observations, effectively leveraging historical context to refine the current understanding of the scene?

    \item \textbf{Feature learning in polar space.} How can we extract general-purpose features in polar coordinates that remain both informative and free from spatial distortions?
\end{enumerate}

Our Polar-Fast-Cartesian-Full (PFCF) (Fig. ~\ref{fig:pipeline}) addresses these challenges through (1) introducing a sector feature buffer for full-scene reasoning (Sec.~\ref{sec:buffer}), (2) Polar Hierarchical Mamba: a Mamba-based streaming backbone specially designed for the polar coordinate system (Sec.~\ref{sec:phim}), and (3) dimensionally-decomposed convolutions to avoid distortion-heavy planes (Sec.~\ref{sec:ddc}).

\vspace{-2mm}
\subsection{Sector Feature Buffer}
\label{sec:buffer}
\vspace{-5pt}

A fundamental weakness of the streaming paradigm lies in the inevitable loss of information when processing partial LiDAR sectors independently. Although implicit hidden states can propagate limited context forward, long-range cross-sector dependencies are weakened with increasing context lengths. Moreover, because predictions are made sector by sector, streaming detectors frequently suffer at sector boundaries. Objects that straddle two angular sectors often produce incomplete foreground evidence in each, leading either to missed detections or duplicate predictions across sectors. Prior works attempt to address these issues with post-processing heuristics such as stateful non-maximum suppression~\cite{han} or initial context padding~\cite{polarstream}, yet these remain insufficient to close the gap between streaming and full-scan detection.

\textbf{Sector Feature Buffer (SFB).} 
We introduce the \textit{Sector Feature Buffer (SFB)} to bridge this gap by unifying polar streaming features with global Cartesian context. The SFB acts as a dynamic, lightweight memory that accumulates sector-level representations produced by the polar streaming backbone. Rather than discarding sector features after local prediction, the buffer incrementally stitches them together into a complete global feature map aligned with the LiDAR sweep. This stitched representation allows each incoming sector to immediately contribute to scene-wide predictions, while maintaining high update frequency.

\textbf{Polar-to-Cartesian Mapping.} 
A central role of the SFB is to transform the stitched polar map into a Cartesian bird’s-eye-view (BEV) representation, enabling global spatial reasoning without waiting for a full scan. Given polar voxel size $(\Delta r, \Delta \theta, \Delta z)$, each sector-level polar voxel $(r_i, \theta_j, z_k)$ is first converted to real-world coordinates:
\begin{equation}\small
\begin{aligned}
r &= r_{\text{offset}} + r_i \cdot \Delta r, 
\quad \theta = \theta_{\text{offset}} + \theta_j \cdot \Delta \theta, \\
x &= r \cdot \cos(\theta), \quad y = r \cdot \sin(\theta), \quad
z = z_{\text{offset}} + z_k \cdot \Delta z.
\end{aligned}
\end{equation}
\vspace{-5pt}

These continuous Cartesian coordinates $(x,y,z)$ are mapped into integer voxel indices within a global Cartesian grid by subtracting the minimum bounds $(x_{\text{offset}}, y_{\text{offset}}, z_{\text{offset}})$ of the Cartesian voxel grid:
\begin{equation}\small
x' = \left\lfloor \frac{x - x_{\text{offset}}}{\Delta x} \right\rfloor, \ \
y' = \left\lfloor \frac{y - y_{\text{offset}}}{\Delta y} \right\rfloor, \ \
z' = \left\lfloor \frac{z - z_{\text{offset}}}{\Delta z} \right\rfloor.
\end{equation}
Only valid indices inside the grid bounds are retained. Each valid feature is then scattered into the dense tensor using:
\begin{equation}
\text{linear\_index} = z' \cdot (N_x \cdot N_y) + y' \cdot N_x + x',
\end{equation}
where $N_x, N_y$ are the spatial dimensions of the Cartesian voxel grid. If multiple features fall into the same bin, they are averaged. After filling the 3D tensor, the vertical axis is collapsed into the channel dimension, yielding a 2D BEV feature map of shape $(N_x, N_y)$. This BEV map can then be processed efficiently with a lightweight Cartesian backbone, enabling explicit cross-sector communication and global reasoning at streaming frequency.

\textbf{Hybrid Integration.} 
Through this design, the SFB naturally integrates the strengths of both paradigms: fast, sector-level polar encoding for responsiveness, and Cartesian BEV reasoning for accuracy and consistency. In effect, the SFB allows our detector to perform \textit{full-scan style global processing at streaming speed}.

\vspace{-8pt}
\subsection{Polar Hierarchical Mamba (PHiM)} 
\label{sec:phim}
\vspace{-3pt}

Streaming LiDAR detection requires mechanisms to capture spatiotemporal dependencies across multiple partial sectors. Prior methods typically apply a standalone model to each sector and add compensatory mechanisms for limited context. For example, Han~\cite{han} uses explicit memory modules across time, STROBE~\cite{strobe} incorporates auxiliary modalities such as HD maps, and PolarStream~\cite{polarstream} applies contextual padding to recover objects at sector boundaries. In contrast, we leverage the Mamba state space model~\cite{mamba} to treat streaming LiDAR as a temporal sequence of egocentric sectors. Rather than relying on external memory or handcrafted fusion, our Polar Hierarchical Mamba (PHiM) maintains temporal information implicitly via the Mamba hidden state, smoothly carrying forward features across sector boundaries. This yields a lightweight yet expressive framework for spatiotemporal modeling without auxiliary modules. For details of the Mamba formulation, please refer to the Appendix \ref{app:mamba}.

Multiresolution feature learning is also essential for competitive performance, as seen in hierarchical downsampling~\cite{SAFDNet, HEDNet}, implicit window embeddings~\cite{voxelmamba}, and attention-based methods~\cite{DSVT}. Inspired by hierarchical designs from other modalities~\cite{swin, localvim}, we propose a streaming architecture that captures intra-sector structure with bidirectional SSM, extracts sector features using dimensionally-decomposed convolutions, and models inter-sector dependencies with forward SSM in the temporal dimension. This design enables fine-grained local details and global spatiotemporal context, while naturally serializing sectors by time order -- removing the need for geometric serialization heuristics such as Hilbert or Z-order curves~\cite{hilbert, zcurve}.

As shown in Fig.~\ref{fig:block_diagram}, sparse input features are first serialized by azimuth, radial distance, and height -- directly from polar voxel indices. Two Mamba blocks process features in forward and backward directions, and their outputs are summed to enrich local spatial representation within each sector. High-level sector features are then extracted using decomposed convolutions (Sec.~\ref{sec:ddc}), avoiding distortion in the $(r, \theta)$ plane. Finally, these features are re-serialized and passed into a global Mamba block, which aggregates cross-sector context through its hidden state. Pseudocode and implementation details are provided in the Appendix.

\vspace{-8pt}
\subsection{Dimensionally-decomposed convolutions}
\label{sec:ddc}
\vspace{-3pt}

Streaming LiDAR methods commonly adopt polar representations to reduce the computational and memory overhead of cuboid voxelization \cite{polarstream}, while better aligning with the anisotropic sparsity of point clouds \cite{partner}. However, this representation introduces spatial distortion, where objects of similar physical size may be drastically warped in the polar view. This distortion undermines the assumption of translation invariance on which convolution-based methods rely, decreasing kernel generalizability. Despite this, convolutions remain prevalent due to their speed, efficiency, and capacity to expand the effective receptive field for modeling long-range dependencies -- an essential trait for handling sparse LiDAR data \cite{SAFDNet}. Consequently, prior polar-based approaches retain convolutional backbones but require substantial post-processing, such as parameterized bilinear sampling \cite{polarstream} or keypoint attention \cite{partner}, to compensate for the accumulated distortion. 

Despite leveraging a convolution-free Mamba backbone (Sec. \ref{sec:phim}) for less inductive bias, we retain convolutions for downsampling as a simple way to capture long-range connections between nonempty regions. Through our evaluation of polar distortion, we find that the \((r, \theta)\) plane accounts for over half the total distortion. Thus, we opt to simply avoid convolving over the \((r, \theta)\) plane and decompose 3D convolutions into two 2D convolutions -- over \((r, z)\) and \((\theta, z)\). For more details in our evaluation process, please refer to Appendix \ref{sec:metrics_eval}.

\begin{figure}[ht]
  \centering
  \includegraphics[width=0.9\linewidth]{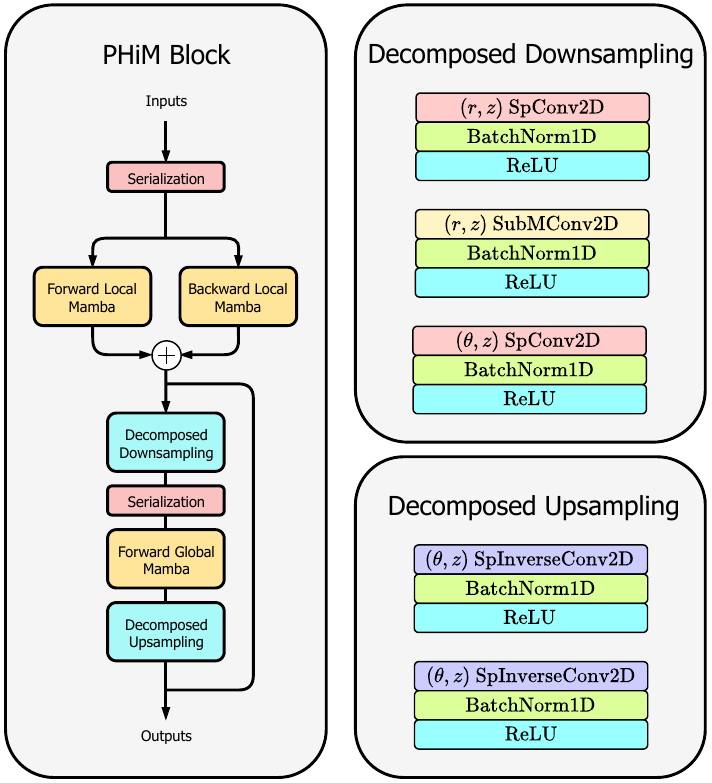}
  \caption{\textbf{PHiM Block.} Serialization is according to the azimuth angle, and the bidirectional local SSM is aggregated with an elementwise addition.}
  \vspace{-2mm}
  \label{fig:block_diagram}
\end{figure}

We present our decomposed downsampling and upsampling strategy in Fig. \ref{fig:block_diagram} (right). Downsampling uses a sparse 2D convolution with stride \((3, 3)\) on \((z, r)\), followed by a submanifold 2D convolution with the same stride, then a sparse convolution with stride \((1, 3)\) on \((z, \theta)\). Upsampling involves inverse convolutions with the same stride parameters but in reverse. The third axis is reshaped into a batch dimension, enabling independent processing. This decouples the scaling effects of \(r\) and \(\theta\), enabling the convolution kernels to generalize better in planes with higher translation invariance.

\vspace{-5pt}
\subsection{Training and implementation}
\label{sec:training}
\vspace{-5pt}

We use the open-souce OpenPCDet \cite{openpcdet} toolbox to implement our method. During training, we simulate a streaming setup by transforming full point clouds into partial sectors, splitting along the azimuth angles. Each sector is treated as an individual batch sample to prevent token-mixing between time steps. This process is illustrated at the bottom left of Fig. \ref{fig:pipeline}. We train on 8 H200 GPUs and evaluate latencies on batch size of 1 on 1 H200 GPU. For sensor latency estimations, we assume a Velodyne HDL-64E scanning LiDAR, which takes about 100 ms for one full 360 degree scan \cite{Waymo}. This is a widely used LiDAR sensor across multiple benchmarks \cite{nuScenes, Waymo} and common to autonomous vehicles. The main implication of this assumption is that all full scan methods would be limited to a maximum throughput of 10 predictions per second, as they must wait for a fully aggregated point cloud before initiating the forward pass.
\begin{table*}[t]
    \centering
    \resizebox{\linewidth}{!}{
    
        \begin{tabular}{c|c|c|c|c|c|c|c|c|c}
        \toprule
        \multirow{2}{*}{\textbf{Type}} & \multirow{2}{*}{\textbf{Method}} & \multicolumn{2}{c|}{\textbf{mAP/mAPH}} & \multicolumn{2}{c|}{\textbf{Vehicle AP/APH}} & \multicolumn{2}{c|}{\textbf{Pedestrian AP/APH}} & \multicolumn{2}{c}{\textbf{Cyclist AP/APH}} \\
        \cmidrule{3-10}
        & & \textbf{L1} & \textbf{L2} & \textbf{L1} & \textbf{L2} & \textbf{L1} & \textbf{L2} & \textbf{L1} & \textbf{L2} \\
        \midrule
        \multirow{8}{*}{Full} & SECOND~\cite{SECOND} & 67.2/63.1 & 61.0/57.2 & 72.3/71.7 & 63.9/63.3 & 68.7/58.2 & 60.7/51.3 & 60.6/59.3 & 58.3/57.0\\
        & PointPillar~\cite{PointPillar} & 69.0/63.5 & 62.8/57.8 & 72.1/71.5 & 63.6/63.1 & 70.6/56.7 & 62.8/50.3 & 64.4/62.3 & 61.9/59.9 \\
        & CenterPoint~\cite{CenterPoint} & 75.9/73.5 & 69.8/67.6 & 76.6/76.0 & 68.9/68.4 & 79.0/73.4 & 71.0/65.8 & 72.1/71.0 & 69.5/68.5 \\
        & DSVT-Voxel~\cite{DSVT} & 80.3/78.2 & 74.0/72.1 & 79.7/79.3 & 71.4/71.0 & 83.7/78.9 & 76.1/71.5 & 77.5/76.5 & 74.6/73.7 \\
        & HEDNet~\cite{HEDNet} & 81.4/79.4 & 75.3/73.4 & 81.1/80.6 & 73.2/72.7 & 84.4/80.0 & 76.8/72.6 & 78.7/77.7 & 75.8/74.9 \\
        & VoxelNeXt~\cite{VoxelNeXt} & 78.6/76.3 & 72.2/70.1 & 78.2/77.7 & 69.9/69.4 & 81.5/76.3 & 73.5/68.6 & 76.1/74.9 & 73.3/72.2 \\
        & Voxel Mamba~\cite{voxelmamba} & -/79.6 & -/73.6 & 80.8/80.3 & 72.6/72.2 & 85.0/80.8 & 77.7/73.6 & 78.6/77.6 & 75.7/74.8 \\
        & UniMamba~\cite{unimamba} & -/- & 76.1/74.1 & 80.6/80.1 & 72.3/71.8 & 86.0/81.3 & 78.7/74.1 & 80.3/79.3 & 77.5/76.5 \\        
        \midrule
        \multirow{4}{*}{Stream} 
        & PolarStream~\cite{polarstream}* & -/- & -/60.8 & 72.4/71.8 & 64.6/64.0 & -/- & -/- & -/- & -/-\\
        & FPA-3DOD~\cite{fastpolarattentive} & -/- & -/- & 76.3/- & 69.8/- & 72.7/- & 70.1/- & -/- & -/-\\
        & PARTNER~\cite{partner} & -/- & -/63.2 & 76.1/75.5 & 68.6/68.1 & -/- & -/- & -/- & -/-\\
        & \textbf{PFCF (ours)} & 78.5/76.6 & 72.1/70.3 & 79.2/78.6 & 71.0/70.5 & 80.7/76.6 & 72.7/68.8 & 75.5/74.4 & 72.7/71.7\\
        
        \bottomrule
        \end{tabular}
    
    }
    \caption{\textbf{Full scan results on Waymo Open validation set.} Metrics: mAP/mAPH(\%)$\uparrow$ for overall results, AP/APH (\%)$\uparrow$ for each category. * denotes reimplementation results. mAP denotes mean average precision and mAPH denotes mean average precision with heading. AP and APH are per-category average precision and average precision with heading. \vspace{-13pt}}
    \label{waymo_results}
\end{table*}

\usepgfplotslibrary{groupplots}
\pgfplotsset{compat=newest}

\begin{figure}[ht]
    \centering
    \resizebox{\linewidth}{!}{
    \begin{tikzpicture}
        \begin{groupplot}[
            group style={
                group size=2 by 1,
                horizontal sep=0.5cm,
                vertical sep=0cm,
                ylabels at=edge left,
                yticklabels at=edge left,
                xticklabels at=edge bottom,
            },
            width=9cm,
            height=8cm,
            ylabel={\textbf{Average Precision (L2 mAPH) [\% $\uparrow$]}},
            xmin=0, xmax=50,
            ymin=55, ymax=76,
            xmode=log,
            log basis x=2,
            grid=major,
            tick label style={font=\large},
            label style={font=\large},
            title style={font=\bfseries\Large},
            legend style={
                font=\large,
                at={(1.0,-0.2)},
                anchor=north,
                legend columns=4,
                /tikz/every even column/.append style={column sep=0.3cm}
            }
        ]

        \nextgroupplot[
            title={\textbf{Throughput vs mAPH}},
            xlabel={\textbf{Throughput [predictions/sec $\uparrow$]}}
        ]
        \addplot[thick, red, dash pattern=on 1pt off 1pt] coordinates {(10,55) (10,76)};
        \addlegendentry{Full-scan max throughput}

        \addplot[only marks, mark=*, mark size=2, mark options={fill=red}] coordinates {
            (9.19, 73.4)
            (4.87, 70.1)
            (10.00, 67.6)
            (10.00, 57.8)
            (10.00, 57.2)
            (8.25, 74.1)
            (10.00, 73.6)
            (10.00, 72.1)
        };
        \addlegendentry{Full-scan methods}

        \addplot[color=cyan, mark=*, mark options={fill=cyan}, mark size=2] coordinates {
            (10.00, 63.1)
            (19.23, 63.9)
            (27.03, 65.3)
            (32.26, 64.8)
            (40.00, 64.0)
        };

        \addplot[color=cyan, mark=*, mark options={fill=cyan}, mark size=2, forget plot] coordinates {
            (10.00, 60.9)
            (19.61, 61.1)
            (27.03, 60.8)
            (34.48, 59.7)
            (41.67, 60.2)
        };
        \addlegendentry{Streaming methods}

        \addplot[
            thick,
            dashed,
            color=black,
            mark=triangle*,
            mark options={fill=yellow},
            mark size=4
        ] coordinates {
            (10.00, 70.3)
            (19.23, 70.1)
            (25.64, 70.3)
            (29.41, 69.7)
            (31.25, 69.0)
        };
        \addlegendentry{Ours}

        \node[anchor= south ] at (axis cs:27.03, 66.3) {\small \cite{partner} };
        \node[anchor= north ] at (axis cs:34.48, 59.7) {\small \cite{polarstream}};
        \node[anchor=north east] at (axis cs:9.19, 73.4) {\small  \cite{HEDNet}};
        \node[anchor=north west] at (axis cs:4.87, 70.1) {\small \cite{VoxelNeXt}};
        \node[anchor=north east] at (axis cs:10.00, 67.6) {\small \cite{CenterPoint}};
        \node[anchor=south east] at (axis cs:10.00, 57.8) {\small \cite{PointPillar}};
        \node[anchor=north west] at (axis cs:10.00, 57.2) {\small \cite{SECOND}};
        \node[anchor= south east] at (axis cs:8.25, 74.1) {\small \cite{unimamba}};
        \node[anchor= south west] at (axis cs:10.00, 73.6) {\small \cite{voxelmamba}};
        \node[anchor= west] at (axis cs:10.00, 72.1) {\small \cite{DSVT}};
        \node[anchor= south] at (axis cs:31.25, 70.7) {\small \textbf{PFCF (Ours)}};

        \node[anchor= south west] at (axis cs:10.00, 63.1) {\small 1};
        \node[anchor= south east] at (axis cs:19.23, 63.9) {\small 1/2};
        \node[anchor= south east] at (axis cs:27.03, 65.3) {\small 1/4};
        \node[anchor= south] at (axis cs:32.26, 64.8) {\small 1/6};
        \node[anchor= south] at (axis cs:40.00, 64.0) {\small 1/8};

        \node[anchor= south ] at (axis cs:10.99, 60.9) {\small 1};
        \node[anchor= south east] at (axis cs:19.61, 61.1) {\small 1/2};
        \node[anchor= south east] at (axis cs:26.31, 60.9) {\small 1/4};
        \node[anchor= south] at (axis cs:34.48, 59.7) {\small 1/6};
        \node[anchor= south] at (axis cs:41.67, 60.2) {\small 1/8};

        \node[anchor= south west] at (axis cs:10.00, 70.3) {\small 1};
        \node[anchor= south east] at (axis cs:19.23, 70.1) {\small 1/2};
        \node[anchor= south east] at (axis cs:25.64, 70.3) {\small 1/4};
        \node[anchor= south west] at (axis cs:29.41, 69.7) {\small 1/6};
        \node[anchor= west] at (axis cs:31.25, 69.0) {\small 1/8};

        \nextgroupplot[
            title={\textbf{Inference Speed vs mAPH}},
            xlabel={\textbf{Inference Speed [frames/sec $\uparrow$]}},
            ylabel={} 
        ]
        \addplot[only marks, mark=*, mark size=2, mark options={fill=red}] coordinates {
            (9.19, 73.4)
            (4.87, 70.1)
            (23.79, 67.6)
            (27.17, 57.8)
            (37.62, 57.2)
            (8.25, 74.1)
            (19.61, 73.6)
            (15.38, 72.1)
        };

        \addplot[color=cyan, mark=*, mark options={fill=cyan}, mark size=2] coordinates {
            (10.64, 63.1)
            (19.23, 63.9)
            (27.03, 65.3)
            (32.26, 64.8)
            (40.00, 64.0)
        };

        \addplot[color=cyan, mark=*, mark options={fill=cyan}, mark size=2] coordinates {
            (10.99, 60.9)
            (19.61, 61.1)
            (27.78, 60.8)
            (34.48, 59.7)
            (43.48, 60.2)
        };

        \addplot[
            thick,
            dashed,
            color=black,
            mark=triangle*,
            mark options={fill=yellow},
            mark size=4
        ] coordinates {
            (12.50, 70.3)
            (19.23, 70.1)
            (25.64, 70.3)
            (29.41, 69.7)
            (31.25, 69.0)
        };

        \node[anchor=  east] at (axis cs:10.64, 63.1) {\scriptsize \cite{partner} };
        \node[anchor=  east ] at (axis cs:10.99, 60.9) {\scriptsize \cite{polarstream}};
        \node[anchor= north west] at (axis cs:9.19, 73.4) {\scriptsize  \cite{HEDNet}};
        \node[anchor= north west] at (axis cs:4.87, 70.1) {\scriptsize \cite{VoxelNeXt}};
        \node[anchor= east] at (axis cs:23.79, 67.6) {\scriptsize \cite{CenterPoint}};
        \node[anchor= east] at (axis cs:27.17, 57.8){\scriptsize \cite{PointPillar}};
        \node[anchor=north east] at (axis cs:37.62, 57.2) {\scriptsize \cite{SECOND}};
        \node[anchor= south east] at (axis cs:8.25, 74.1) {\scriptsize \cite{unimamba}};
        \node[anchor= south west] at (axis cs:19.61, 73.6) {\scriptsize \cite{voxelmamba}};
        \node[anchor= north west] at (axis cs:15.38, 72.1) {\scriptsize \cite{DSVT}};
        \node[anchor= south] at (axis cs:31.25, 70.7) {\scriptsize \textbf{PFCF (Ours)}};

        \node[anchor= south west] at (axis cs:10.00, 63.1) {\tiny 1};
        \node[anchor= south east] at (axis cs:19.23, 63.9) {\tiny 1/2};
        \node[anchor= south east] at (axis cs:27.03, 65.3) {\tiny 1/4};
        \node[anchor= south] at (axis cs:32.26, 64.8) {\tiny 1/6};
        \node[anchor= south] at (axis cs:40.00, 64.0) {\tiny 1/8};

        \node[anchor= south ] at (axis cs:10.99, 60.9) {\tiny 1};
        \node[anchor= south east] at (axis cs:19.61, 61.1) {\tiny 1/2};
        \node[anchor= south east] at (axis cs:26.31, 60.9) {\tiny 1/4};
        \node[anchor= south] at (axis cs:34.48, 59.7) {\tiny 1/6};
        \node[anchor= south] at (axis cs:41.67, 60.2) {\tiny 1/8};
        
        \node[anchor= south west] at (axis cs:12.50, 70.3) {\tiny 1};
        \node[anchor= south east] at (axis cs:19.23, 70.1) {\tiny 1/2};
        \node[anchor= south east] at (axis cs:25.64, 70.3) {\tiny 1/4};
        \node[anchor= south west] at (axis cs:29.41, 69.7) {\tiny 1/6};
        \node[anchor= west] at (axis cs:31.25, 69.0) {\tiny 1/8};

        \end{groupplot}
    \end{tikzpicture}}
    \caption{\textbf{Performance and speed comparison on Waymo Open.} PFCF offers the speed and throughput benefits of streaming methods with the competitive performance of full-scan methods. For streaming methods, $\frac{1}{N}$ denotes the size of each partial sector -- for example, $\frac{1}{4}$ means each sector is one quarter of a full point cloud. \textbf{(Left)} Throughput is measured end-to-end including sensor latency assuming simultaneous sensing and perception. \textbf{(Right)} Inference speed is measured end-to-end without sensor latency on batch size 1.}
    \vspace{-6mm}
    \label{fig:fps_scatterplot}
\end{figure}

\begin{table}[h!]
    \centering
    \resizebox{\linewidth}{!}{ 
        \begin{tabular}{l|l|ccccc} 
        \toprule
        \multirow{2}{*}{\textbf{Method}} & \multirow{2}{*}{\textbf{Rep.}} & \multicolumn{5}{c}{\textbf{Number of streaming sectors}} \\
        \cmidrule{3-7}
        & & \textbf{1} & \textbf{2} & \textbf{4} & \textbf{6} & \textbf{8} \\
        \midrule
        STROBE~\cite{strobe} & Cartesian & 60.5/59.8 & 59.5/58.9 & 58.8/58.3 & 58.3/57.6 & 58.0/57.3 \\
        Han ~\cite{han} & Cartesian & 61.8/61.4 & 61.7/61.1 & 60.7/60.2 & 60.0/59.3 & 59.9/59.3 \\
        PolarStream ~\cite{polarstream} & Polar & 61.4/60.8 & 61.8/61.2 & 61.2/60.7 & 60.3/59.7 & 60.7/60.2 \\
        PARTNER ~\cite{partner} & Polar & 63.8/63.2 & 64.3/63.8 & 66.0/65.5 & 65.3/64.7 & 64.5/64.0 \\
        \textbf{PFCF (Ours)} & Hybrid & \textbf{72.1/70.3} & \textbf{71.2/70.1} & \textbf{70.6/70.3} & \textbf{70.1/69.7} & \textbf{69.5/69.0} \\
        \bottomrule
        \end{tabular}
    }
    \caption{\textbf{Streaming sector performance.} PFCF outperforms existing streaming methods across all number of streaming sectors and sector sizes. We report L2 results.}
    \vspace{-4mm}
    \label{sector_ablation}
\end{table}

\begin{table}[h!]
    \centering
    \resizebox{\linewidth}{!}{ 
        \begin{tabular}{lcccc|c}
        \toprule
        \textbf{Streaming} & \textbf{DDC} & \textbf{PHiM} & \textbf{SFB} & \textbf{BSSM} & \textbf{L1 mAP} \\
        \midrule
        \textcolor{red}{Incompatible} & \textcolor{red}{\xmark} & \textcolor{red}{\xmark} & \textcolor{red}{\xmark} & \textcolor{mygreen}{\checkmark} & 58.02 \\
        \textcolor{red}{Incompatible} & \textcolor{mygreen}{\checkmark} & \textcolor{red}{\xmark} & \textcolor{red}{\xmark} & \textcolor{mygreen}{\checkmark} & 62.61 \\
        \textcolor{red}{Incompatible} & \textcolor{mygreen}{\checkmark} & \textcolor{red}{\xmark} & \textcolor{mygreen}{\checkmark} & \textcolor{mygreen}{\checkmark} & 67.45 \\
        \textcolor{mygreen}{Compatible} & \textcolor{red}{\xmark} & \textcolor{mygreen}{\checkmark} & \textcolor{red}{\xmark} & \textcolor{mygreen}{\checkmark} & 66.31 \\
        \textcolor{mygreen}{Compatible} & \textcolor{red}{\xmark} & \textcolor{mygreen}{\checkmark} & \textcolor{mygreen}{\checkmark} & \textcolor{mygreen}{\checkmark} & 69.71 \\
        \textcolor{mygreen}{Compatible} & \textcolor{mygreen}{\checkmark} & \textcolor{mygreen}{\checkmark} & \textcolor{red}{\xmark} & \textcolor{mygreen}{\checkmark} & 69.24 \\
        \textcolor{mygreen}{Compatible} & \textcolor{mygreen}{\checkmark} & \textcolor{mygreen}{\checkmark} & \textcolor{mygreen}{\checkmark} & \textcolor{red}{\xmark} & 74.92 \\
        \textcolor{mygreen}{Compatible} & \textcolor{mygreen}{\checkmark} & \textcolor{mygreen}{\checkmark} & \textcolor{mygreen}{\checkmark} & \textcolor{mygreen}{\checkmark} & \textbf{78.50} \\
        \bottomrule
        \end{tabular}
    }
    \caption{\textbf{Component by component ablation.} Waymo L1 mAP for combinations of decomposed depthwise convolutions (DDC), Polar Hierarchical Mamba (PHiM), sector feature buffer (SFB), and bidirectional state-space models (BSSM). Each component contributes to the final performance.}
    \label{ablation}\vspace{-15pt}
\end{table}

\begin{table*}[t]
    \centering
    \resizebox{\linewidth}{!}{
    
        \begin{tabular}{c|c|c|c|c|c|c|c|c|c|c|c|c|c}
        \toprule
        \textbf{Type} & \textbf{Method} & \textbf{NDS} & \textbf{mAP} & \textbf{Car} & \textbf{Truck} & \textbf{Bus} & \textbf{T.L.} & \textbf{C.V.} & \textbf{Ped.} & \textbf{M.T.} & \textbf{Bike} & \textbf{T.C.} & \textbf{B.R.} \\
        \midrule
        \multirow{7}{*}{Full} 
        & PointPillar~\cite{PointPillar} & 46.8 & 28.2 & 75.5 & 31.6 & 44.9 & 23.7 & 4.0 & 49.6 & 14.6 & 0.4 & 8.0 & 30.0 \\
        & CenterPoint~\cite{CenterPoint} & 66.5 & 59.2 & 84.9 & 57.4 & 70.7 & 38.1 & 16.9 & 85.1 & 59.0 & 42.0 & 69.8 & 68.3 \\
        & DSVT-Voxel~\cite{DSVT} & 71.1 & 66.4 & 87.4 & 62.6 & 75.9 & 42.1 & 25.3 & 88.2 & 74.8 & 58.7 & 77.8 & 70.9 \\
        & HEDNet~\cite{HEDNet} & 71.4 & 66.4 & 87.4 & 62.6 & 75.9 & 42.1 & 25.3 & 88.2 & 74.8 & 58.7 & 77.8 & 70.9 \\
        & VoxelNeXt~\cite{VoxelNeXt} & 66.7 & 60.5 & 83.9 & 55.5 & 70.5 & 38.1 & 21.1 & 84.6 & 62.8 & 50.0 & 69.4 & 69.4 \\
        & Voxel Mamba~\cite{voxelmamba} & 71.9 & 67.5 & 87.9 & 62.8 & 76.8 & 45.9 & 24.9 & 89.3 & 77.1 & 58.6 & 80.1 & 71.5 \\
        & UniMamba~\cite{unimamba} & 72.6 & 68.5 & 88.7 & 64.7 & 79.7 & 47.9 & 28.7 & 89.7 & 74.6 & 59.1 & 79.5 & 72.3 \\
        \midrule
        \multirow{2}{*}{Stream} 
        & PolarStream~\cite{polarstream} & 60.0 & 51.2 & 80.4 & 42.5 & 53.5 & 29.8 & 15.9 & 79.8 & 58.0 & 28.5 & 61.3 & 62.3\\
        & \textbf{PFCF (ours)} & 68.5 & 64.2 & 85.0 & 60.6 & 74.8 & 41.7 & 25.1 & 85.1 & 71.8 & 56.6 & 74.3 & 66.8 \\
        
        \bottomrule
        \end{tabular}
    
    }
    \caption{\textbf{Full scan results on nuScenes validation set.} Metrics: mAP(\%)$\uparrow$ for overall results and each individual category, NDS (\%)$\uparrow$ for overall results. mAP denotes mean average precision and NDS denotes nuScenes detection score. \vspace{-10pt}}
    \label{nuscenes_results}
\end{table*}

\begin{table*}[t]
    \centering
    \resizebox{\linewidth}{!}{
    
        \begin{tabular}{c|c|c|c|c|c|c|c|c}
        \toprule
        \multirow{2}{*}{\textbf{Method}} & \multicolumn{2}{c|}{\textbf{mAP/mAPH}} & \multicolumn{2}{c|}{\textbf{Vehicle AP/APH}} & \multicolumn{2}{c|}{\textbf{Pedestrian AP/APH}} & \multicolumn{2}{c}{\textbf{Cyclist AP/APH}} \\
        \cmidrule{2-9}
        & \textbf{L1} & \textbf{L2} & \textbf{L1} & \textbf{L2} & \textbf{L1} & \textbf{L2} & \textbf{L1} & \textbf{L2} \\
        \midrule

        Voxel Mamba 1/4 ~\cite{voxelmamba} & 24.6/23.0 & 22.2/20.8 & 24.9/24.6 & 21.6/21.3 & 23.7/20.3 & 20.4/17.4 & 25.2/24.3 & 24.6/23.7 \\
        \textbf{PHiM 1/4 (ours)} & 40.5/36.5 & 36.4/32.8 & 46.2/45.5 & 39.9/39.3 & 35.1/25.9 & 30.1/22.2 & 40.2/38.0 & 39.2/37.1\\
        
        \bottomrule
        \end{tabular}
    
    }
    \caption{\textbf{Streaming comparison with Mamba baseline on Waymo 1/100.} Metrics: mAP/mAPH(\%)$\uparrow$ for overall results, AP/APH (\%)$\uparrow$ for each category. We report averaged results across 10 separate 1/100 subsamples of Waymo Open validation set and compare methods using sector size of $\frac{1}{4}$. Our contributions boost the spatiotemporal modeling ability of regular Mamba and removes distortion heavy 3D convolutions for superior performance on the polar streaming setting. \vspace{-15pt}}
    \label{mamba_ablation_results}
\end{table*}
\vspace{-4pt}
\section{Results}
\vspace{-8pt}

We structure our results section to answer several main questions.

\begin{enumerate}
    \item How is performance and speed evaluated, especially when comparing streaming and full-scan methods? 

    \item How does PFCF compare to existing streaming and full-scan methods, in terms of performance and update rate? 

    \item How does PFCF performance scale with respect to different streaming granularities (sector sizes)?

    \item How does PHiM compare to baseline Mamba architectures on the streaming setting? Is PHiM truly necessary for good performance?

    \item What is the impact of each component within PFCF towards overall performance?
\end{enumerate}

In addition to these questions, we also provide a thorough analysis of polar distortion propagation, accuracy and efficiency comparison against a Mamba baseline, and evaluation of prediction refinement in the Appendix. 

\vspace{-8pt}
\paragraph{Measuring speed.} We report two speed-related metrics -- latency and throughput -- in order to highlight the main benefits of streaming methods and PFCF. We measure latency similarly to prior literature \cite{DSVT}, recording the mean end-to-end CUDA walltimes over 1000 samples, excluding sensor latency. We measure throughput as the number of predictions per second assuming simultaneous operation of the LiDAR sensor and perception algorithm. Throughput is measured including the overall sensor latency. Since full-scan methods must wait for a full point cloud to be aggregated, the theoretical minimum time it takes from sensing to perception would be 100 ms, and therefore the maximum throughput for full-scan methods is 10 predictions per second. We measure the speed metrics using a single H200 GPU with a batch size of 1. For full-scan methods, a batch size of 1 corresponds to 1 full point cloud. For streaming methods using sector size $\frac{1}{N}$, a batch size of 1 corresponds to a partial sector of size $\frac{1}{N}$.

\vspace{-8pt}
\paragraph{Measuring accuracy.} We ensure fair comparison between PFCF, streaming, and full-scan methods by standardizing the amount of information available to each model. For streaming methods such as PARTNER \cite{partner}, this involves aggregating predictions across multiple partial sectors until they collectively represent one full point cloud. For example, if each partial sector corresponds to one-quarter of the scan, we evaluate the model over four sectors and then combine their predictions. For PFCF, we pass the sequence of partial sectors as input and evaluate just the predictions associated with the last sector.


\vspace{-12pt}
\paragraph{Accuracy results on Waymo Open.} Table ~\ref{waymo_results} shows a comparison of PFCF with existing state of the art full-scan and streaming methods on the Waymo Open \cite{Waymo} validation set. Notably, our method presents a 10\% performance increase in mean average precision (mAP) over the previous leading streaming method PARTNER \cite{partner} and showcases competitive performance against leading full-scan methods. Previous streaming methods only report results on the Vehicle class but we provide results of PFCF on all classes, which is why the Pedestrian and Cyclist classes are empty. 

\vspace{-12pt}
\paragraph{Speed and throughput evaluation on Waymo Open.} We demonstrate the benefits of streaming methods by providing a side-by-side comparison of latency and overall throughput.  As shown in Fig.  ~\ref{fig:fps_scatterplot}, full-scan methods are capped at a theoretical maximum throughput of 10 predictions per second on account of the 100 ms sensor latency, whereas streaming methods are not subject to this full-scan latency. For example, streaming algorithms operating on $\frac{1}{4}$ sectors can initiate four forward passes in the time of a single full scan, essentially unlocking the potential for 4x higher reactiveness. For PFCF, we observe a speed increase with decreasing sector sizes, which is a result of shorter sequences being passed into the SSM components of the PHiM block. We observe slightly worse inference speed and throughput scaling with decreasing sector sizes compared to previous streaming methods, which we attribute to the Mamba blocks having more overhead and being less optimized due to recency. Additionally, we notice slower inference speed on our $\frac{1}{1}$ model compared to a similar full-scan Cartesian model \cite{voxelmamba}, which is a result of our decomposition of 3D convolutions into two separate 2D convolutions increasing the number of sequential layers within the 3D backbone.

\vspace{-15pt}
\paragraph{Sector granularity comparison.} We include evaluations for streaming methods on $\frac{1}{1}, \frac{1}{2}, \frac{1}{4}, \frac{1}{6} \text{and} \frac{1}{8}$ sector sizes in Table ~\ref{sector_ablation}. These streaming sizes are typical choices for streaming methods as even smaller sector sizes would no longer be useful since the algorithm latency would become the new bottleneck. The sector size reflects how many individual sectors a full point cloud is split into (see Fig. \ref{fig:overview}). Overall, PFCF far outperforms previous streaming methods in precision while offering the streaming benefit of higher throughput and inference speed with decreasing sector sizes. 

\vspace{-15pt}
\paragraph{Component-by-component ablation.} Table ~\ref{ablation} ablates our decomposed convolutions (DDC), Polar Hierarchical Mamba (PHiM) streaming backbone, sector feature buffer, and bidirectional SSM within PHiM. We report the $L_1$ mean average precision on the validation set. Crucially, we find that all components are necessary for maximum performance: decomposed convolutions to capture spatial information without distortion warp, PHiM to propagate inter-sector information, the bidirectional SSM to fully extract intra-sector features, and the sector feature buffer to enable full scene understanding.

\vspace{-10pt}
\paragraph{Baseline Mamba comparison.} Table ~\ref{mamba_ablation_results} provides a comparison of our Polar Hierarchical Mamba against the Voxel Mamba architecture. We tweak Voxel Mamba to support the streaming setting and use it as a drop in replacement for PHiM within the PFCF pipeline. We observe that PHiM far outperforms Voxel Mamba on the streaming setting, which we attribute to an overall more compatible architecture for the polar setting. Voxel Mamba contains windowed positional embeddings and serialization curves that mimic convolutional assumptions, which are actually detrimental when applied to the streaming setting.

\vspace{-8pt}
\paragraph{Results on nuScenes.} We report the accuracy of PFCF on the nuScenes validation set in Table ~\ref{nuscenes_results}. Overall, performance is relatively stable for PFCF across both datasets, and PFCF still demonstrates superior performance compared to other streaming methods \cite{polarstream}.
\vspace{-5pt}
\section{Conclusion}
\vspace{-5pt}
We introduce Polar-Fast-Cartesian-Full (PFCF), a novel detector designed for reactive and accurate LiDAR perception. By first processing partial sectors independently with a fast Polar Hierarchical Mamba (PHiM) backbone, stitching together sector features to construct a full scene feature map and capturing global scene understanding with an accurate and lightweight 2D Cartesian backbone, PFCF enables full-scan processing at streaming speeds. Evaluated on the Waymo Open Dataset, PFCF sets a new state-of-the-art among streaming models and rivals Cartesian full-scan baselines in accuracy while delivering twice the throughput. 

\section{Acknowledgement}
\thanks{This work was supported in part by CogniSense, one of seven centers in JUMP 2.0, a Semiconductor Research Corporation (SRC) program sponsored by DARPA. This research was also supported through research cyberinfrastructure resources and services provided by the Partnership for an Advanced Computing Environment (PACE) at the Georgia Institute of Technology, Atlanta, Georgia, USA.}

{
    \small
    \bibliographystyle{ieeenat_fullname}
    \bibliography{main}
}

\clearpage
\appendix
\setcounter{page}{1}
\maketitle

\section{Mamba}\label{app:mamba}
\subsection{Background}
The state space (SSM) model \cite{S4} continuous system maps a 1D input $x(t) \in \mathbb{R}^L$ to an output signal $y(t) \in \mathbb{R}^L$ via a hidden state $h(t) \in \mathbb{R}^N$. This can be represented as the following set of linear differential equations:

\vspace{-9pt}
\begin{equation}
\left\{
\begin{aligned}
    h'(t) &= \mathbf{A}h(t) + \mathbf{B}x(t), \\
    y(t) &= \mathbf{C}h(t) + \mathbf{D}x(t),
\end{aligned}
\right.
\end{equation}
\vspace{-5pt}

where $\mathbf{A} \in \mathbb{R}^{N \times N}$, $\mathbf{B} \in \mathbb{R}^{N \times N}$, $\mathbf{C} \in \mathbb{R}^{1 \times N}$ are the learnable parameters and $\mathbf{D \in \mathbb{R}^1}$ is a residual connection. Mamba \cite{mamba} discretizes the SSM model parameters $\mathbf{A}$ and $\mathbf{B}$ using the zero-order hold (ZOH) transformation and a timescale parameter $\Delta$, where $\overline{\mathbf{A}} = \text{exp}(\mathbf{\Delta A})$, $\overline{\mathbf{B}} = (\mathbf{\Delta A})^{-1}(\text{exp}(\mathbf{\Delta A}) - \mathbf{I}) \cdot \mathbf{\Delta A}$. The resulting discretized equations are as follows:

\vspace{-5pt}
\begin{equation}
\left\{
\begin{aligned}
    h_{t} &= \mathbf{A}h_{t-1} + \mathbf{B}x_{t}, \\
    y_{t} &= \mathbf{C}h_{t},
\end{aligned}
\right.
\end{equation}

where $\mathbf{D}$ is omitted from the equations and incorporated as a simple residual connection in the architecture. Finally, Mamba computes its output through an efficient reformulation as a global convolution, enabling efficient training:

\vspace{-5pt}
\begin{equation}
\left\{
\begin{aligned}
    \overline{\mathbf{K}} &= (\mathbf{C\overline{B}}, \mathbf{C\overline{AB}}, ..., \mathbf{C\overline{A}^{k}\overline{B}}), \\
    \mathbf{y} &= \mathbf{x} * \overline{\mathbf{K}
    }
\end{aligned}
\right.
\end{equation}

Mamba combines the time-varying strength of self-attention, near-linear scaling of recurrent neural networks and fast training of convolutions for efficient modeling of sequences. This enables potential benefits in long-horizon (ie. hundreds of partial sectors across multiple full rotations of the LiDAR sensor) driving scenarios, although we defer this exploration to future work.

\subsection{Comparison of PFCF against existing Mamba works.}\label{app:mamba_comparison}
\begin{figure}

    \centering
    \includegraphics[width=0.9\linewidth]{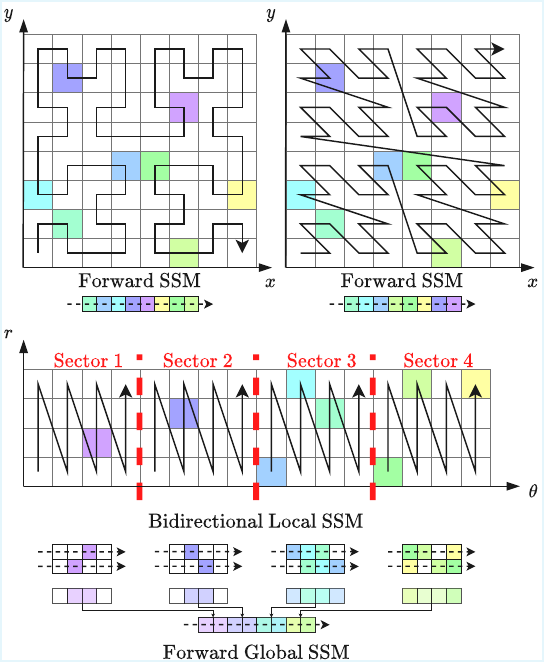}
    \caption{\textbf{Comparison with existing LiDAR Mamba works. (Top)} Previous Mamba-based LiDAR models operate on full point cloud scans and use costly geometric heuristics such as Hilbert (left) or Z curves (right) to maintain spatial locality. \textbf{(Bottom)} Our method serializes sectors simply by when they arrive, spreading memory usage and computations over the course of a sensor rotation rather than altogether in one peak.} 
    \label{fig:mamba_comparison}
\end{figure}

Mamba methods for LiDAR object detection include Voxel Mamba \cite{voxelmamba} and UniMamba \cite{unimamba}. These methods use costly geometric heuristics such as the Hilbert curve or Z-curve to impose an arbitrary order by which the voxel grid is serialized. These heuristics require recomputation for every resolution size as the curve patterns are only square-shaped. This leads to significant memory usage during inference. In comparison, our Polar Hierarchical Mamba backbone offers a simple serialization method by leveraging the natural temporal ordering of the polar space. We simply serialize according to the azimuth angles, using a bidrectional local SSM to fully capture the intra-sector relationships before employing a forward global SSM to capture the inter-sector dependencies. A visualization of the different serialization approaches is shown in Figure ~\ref{fig:mamba_comparison}.

\section{Implementation and training details}

\paragraph{Waymo Open dataset.}
The Waymo Open dataset \cite{Waymo} has a detection range of 75 meters and adopts mean average precision (mAP) and mean average precision weighted by heading accuracy (mAPH) as the main evaluation metrics. Waymo differentiates the metrics into L1 and L2 difficulty levels - L1 for objects with more than five LiDAR points and L2 for objects with at least one LiDAR point. We use a voxel size of $(0.32m, 0.32m, 0.1875m)$ and detection range of $[-74.88m, 74.88m]$ along the X and Y axes. We set the detection range to $[-2m, 4m]$ in the Z axis. We use the Adam \cite{Adam} optimizer with a one-cycle learning rate policy, a weight decay of $0.05$, and max learning rate of $2.5e-3$. We use a class-specific non-max-suppression during inference with IoU thresholds of $0.75, 0.6, 0.55$, corresponding to vehicle, pedestrian, and cyclist classes respectively. A faded training strategy is used for the last epoch.

\paragraph{nuScenes dataset.}
The nuScenes dataset \cite{nuScenes} features a detection range of 54 meters along the X and Y axes, with a vertical range of $[-5m, 3m]$, and uses mean average precision and nuScenes detection score (NDS) averaged across all classes as the main metrics. mAP is averaged across the distance thresholds of $0.5m, 1.0m, 2.0m,$ and $4.0m$. NDS is an average of mAP along with true positive evaluations of translation, velocity, orientation, scaling and attribute errors. We follow the official 10-class setting, including vehicles, barriers, and traffic participants. Our voxelization uses a uniform voxel size of $(0.25m, 0.25m, 0.25m)$ and detection range of $[-54m, 54m]$ along the X and Y axes. We use the Adam optimizer with a one-cycle learning rate schedule, setting a maximum learning rate of $1.5e-3$, a weight decay of $0.01$, and gradient clipping of $10$.

\paragraph{Model implementation.} We use the open-source OpenPCDet \cite{openpcdet} codebase to build our model. For our feature encoder we tweak the dynamic \cite{DynamicVFE} VFE to support cylindrical coordinates, calculating cylindrical voxel sizes to output a voxel grid of shape $(468, 468, 32)$ along the $(r, \theta, z)$ axes respectively. We build our own 3D backbone using the Mamba code provided by \cite{mamba, voxelmamba}, and use the 2D backbone from \cite{PointPillar} and the detection head from \cite{CenterPoint}. To maintain comparability with previous methods we use the same training schedule of 24 epochs for Waymo Open \cite{Waymo} and 30 epochs for nuScenes \cite{nuScenes}. Our training on the full Waymo and nuScenes datasets are conducted on 8 H200 GPUs with total batch size of 24 and 16 respectively, and our ablation studies are conducted on 1 H200 GPU with batch size 3 and 2 respectively. Additionally, pseudocode of the forward pass of the PHiM block is provided in Alg. \ref{alg:phim}, and actual code implementations for PHiM, the sector feature buffer and our streaming training data simulation has been attached in the code supplement.

\paragraph{Hyperparameters.} 
For our 3D backbone, we use an input model dimension of 128 and tensor shape of $(468, 468)$. We use 6 PHiM blocks with strides $[1, 1, 1, 2, 1, 4]$ and kernel size $[3, 3, 3, 3, 3, 5]$ for the decomposed downsampling layers. We keep hyperparameters for the 2D backbone and detection head the same as \cite{voxelmamba}.

\begin{figure}[t]
    \centering
    \includegraphics[width=\linewidth]{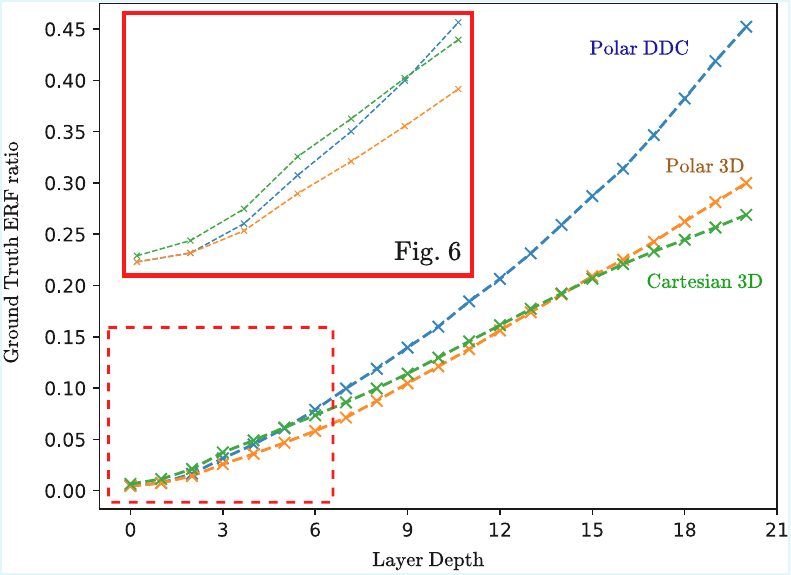}
    \caption{\textbf{Quantitative Distortion over 20 layers.} The physical volume within the effective receptive field (ERF) of ground truth objects compared to background. Over very deep upsampling/downsampling pipelines - for example stacked PHiM blocks - polar decomposed convolutions scale poorly in tracking the foreground ERF ratio of Cartesian 3D convolutions. However, with fewer stacked blocks, polar decomposed convolutions track the Cartesian 3D convolutions ERF better.}
    \label{fig:distortion_full}
\end{figure}
\begin{table*}[t]
    \centering
    \resizebox{\linewidth}{!}{
        \begin{tabular}{l|c|c|c|c}
            \toprule
            \textbf{Method} & \textbf{Num. Params} & \textbf{Params Mem.} & \textbf{FLOPs} & \textbf{Peak Mem. Usage} \\
            \midrule
            Voxel Mamba ~\cite{voxelmamba} & 20.1 M & 76.6 MB & 939.0 G & 3.0 GB \\
            \textbf{PHiM (ours)} & 11.8 M & 45.0 MB & 926.1 G & 1.7 GB \\            
            \bottomrule
        \end{tabular}
    }
    \caption{\textbf{Efficiency comparison with baseline Mamba.} Metrics: Number of parameters in millions ($\downarrow$), parameter memory usage in MB ($\downarrow$), floating point operations in GFLOPs ($\downarrow$), peak memory usage in GB ($\downarrow$). Another benefit of decomposing distortion heavy 3D convolutions into separate 2D convolutions is a reduction in parameters. Besides, PHiM also removes position encodings and serialization curves stored in memory, resulting in significant parameters and memory savings. }
    \label{efficiency}
\end{table*}

\begin{figure}[ht]
    \centering
    \includegraphics[width=\linewidth]{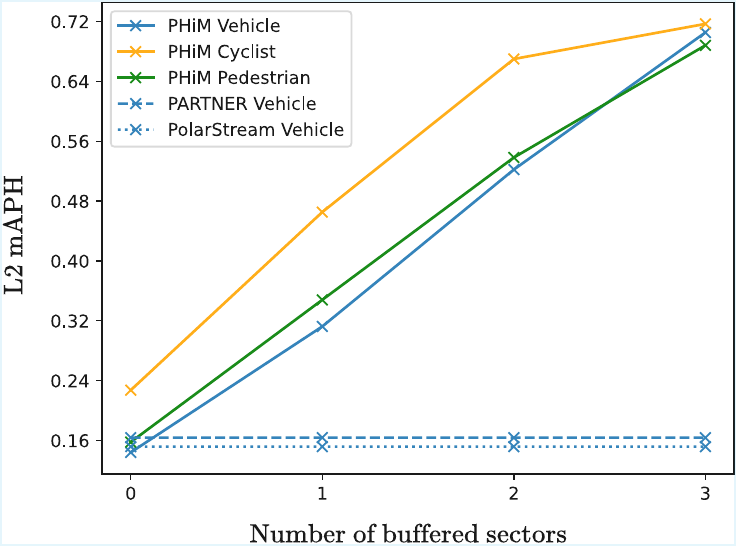}
    \caption{\textbf{Global prediction accuracy over number of buffered sectors.} Global scene prediction performance in relation to number of buffered sector features, evaluated with PHiM $\tfrac{1}{4}$. A buffer size of three therefore corresponds to having full-scan information available (3 buffered sectors and the new input sector). PHiM reformulates streaming prediction to generate and refine global outputs as more sectors are buffered. Unlike other streaming methods that predict only on the current sector without updating past outputs, PHiM progressively improves global accuracy and reaches full-scan performance while retaining streaming efficiency. We report the L2 mAPH for each class individually.}
    \label{fig:refinement_line}
\end{figure}

\begin{figure*}[t]
    \centering
    \includegraphics[width=\textwidth]{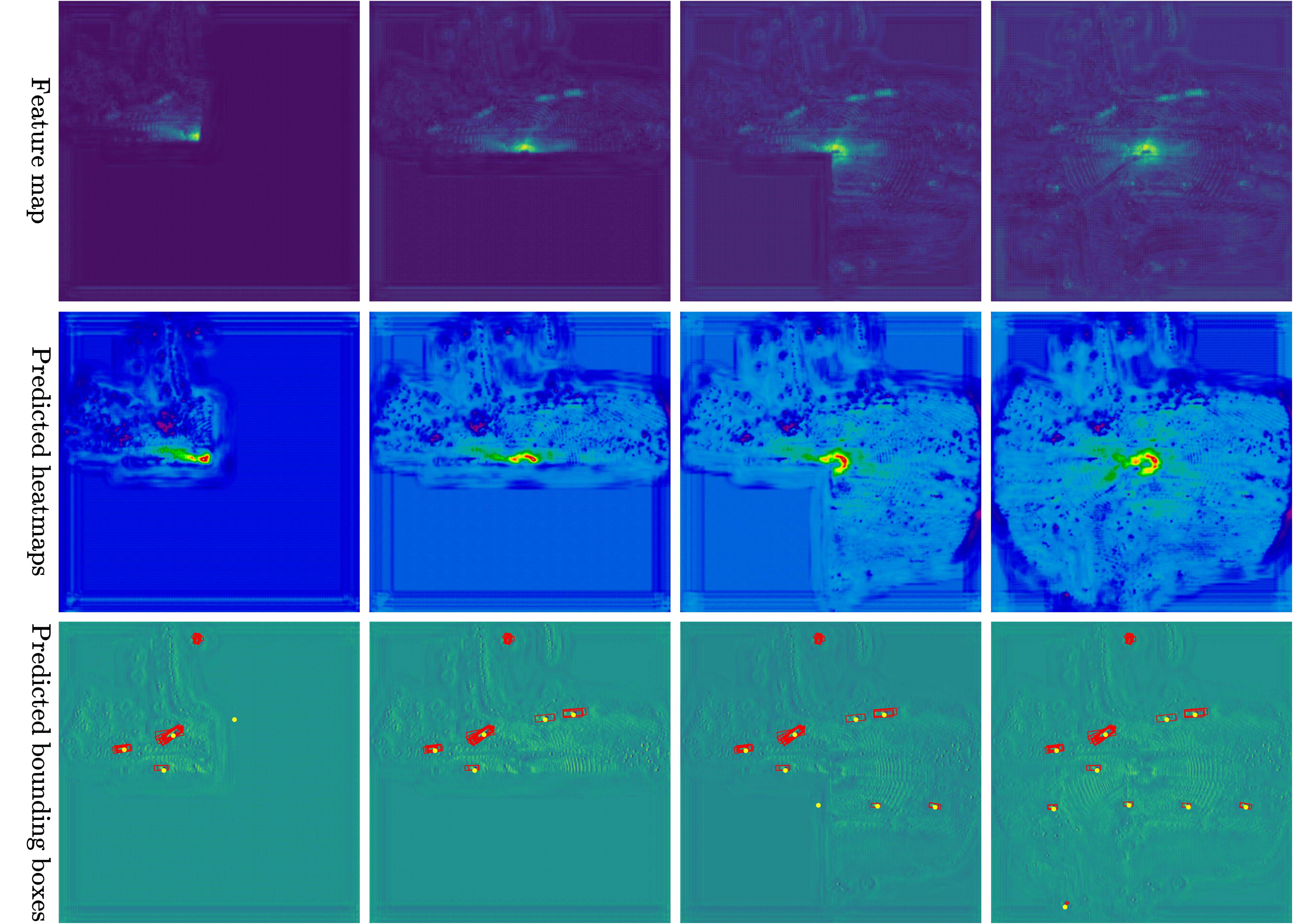}
    \caption{\textbf{Prediction refinement qualitative visualization. (Top)}  PHiM feature map before detection head. From left to right, the frames correspond to the number of sectors in the sector buffer with PHiM operating on $\frac{1}{4}$ sector size. \textbf{(Middle)} Predicted Centerpoint heatmap. \textbf{(Bottom)} Predicted bounding boxes are shown in red and the yellow points indicate ground truth object centers. Ground truths on the sector borders are missed initially (first and third frames) but are able to be detected with information from the following sector (second and fourth frames).}
    \label{fig:refine_qualitative}
\end{figure*}
\begin{algorithm}[ht]
  \caption{\textbf{PHiM Block forward pass.}}
  \label{alg:phim}
  \scriptsize
  \begin{algorithmic}
    \State \textbf{Input:} $\text{Voxel features }V, \text{coordinates } C$ 
    \State $\quad \text{batch size }B, \text{spatial shape }S$
    \State Initialize local feature tensor $V_{\text{local}}$ \\
    \ForEach{frustum $f$}{
      \State extract $V_f$ from $V$
      \State $F_{\text{fw}}\gets \mathrm{fsm}(V_f)$
      \State $F_{\text{bw}}\gets \mathrm{bsm}(V_f^\mathrm{rev})$
      \State $V_{\text{local}}[f] \gets F_{\text{fw}}+F_{\text{bw}}^\mathrm{rev}$
    }
    \State build SparseConvTensor $X\gets(V_{\text{local}},C)$ \\
    \ForEach{encoder $E$}{
      \State $X\gets E(X)$
    }
    \State initialize global feature tensor $V_{\text{global}}$ 
    \State merge partial sectors into same batch \\
    \ForEach{batch $b$}{
      \State extract $F_b$ from $X$
      \State $V_{\text{global}}[b] \gets \mathrm{fgm}(F_b)$
    }
    \State $X.\mathrm{features}\gets\mathrm{normalize}(V_{\text{global}})$ \\
    \ForEach{decoder $D$}{
      \State $X\gets D(X)$
    }
    \State $X.\mathrm{features} \gets X.\mathrm{features} + V_{\text{local}}$
    \State \Return{$X.\mathrm{features},\,X.\mathrm{indices}$}
  \end{algorithmic}
\end{algorithm}

\section{Metrics evaluation details}
\label{sec:metrics_eval}

\paragraph{Comparison with previous methods.}
We use reported accuracy results from original papers in our table and scatterplot. For our latency evaluations, we use specified config files from the OpenPCDet \cite{openpcdet} codebase for the Cartesian based methods and from each repo for the polar-based methods. Due to needing data transformations to simulate a streaming setup, all mAP results are on the validation sets. For comparisons with Cartesian based methods, we evaluate our model with length $1$ sequence of $\frac{1}{1}$ LiDAR sector - equivalent to one full point cloud. 

\paragraph{Ablations.} For our ablations in Tab. ~\ref{ablation} in the main paper, Fig. ~\ref{fig:refinement_line} and Tab. ~\ref{mamba_ablation_results} in the Appendix, we use sector size of $\frac{1}{4}$ for simplicity although any sector size can be chosen for the experiments.

\begin{figure}[ht]
    \centering
    \includegraphics[width=\linewidth]{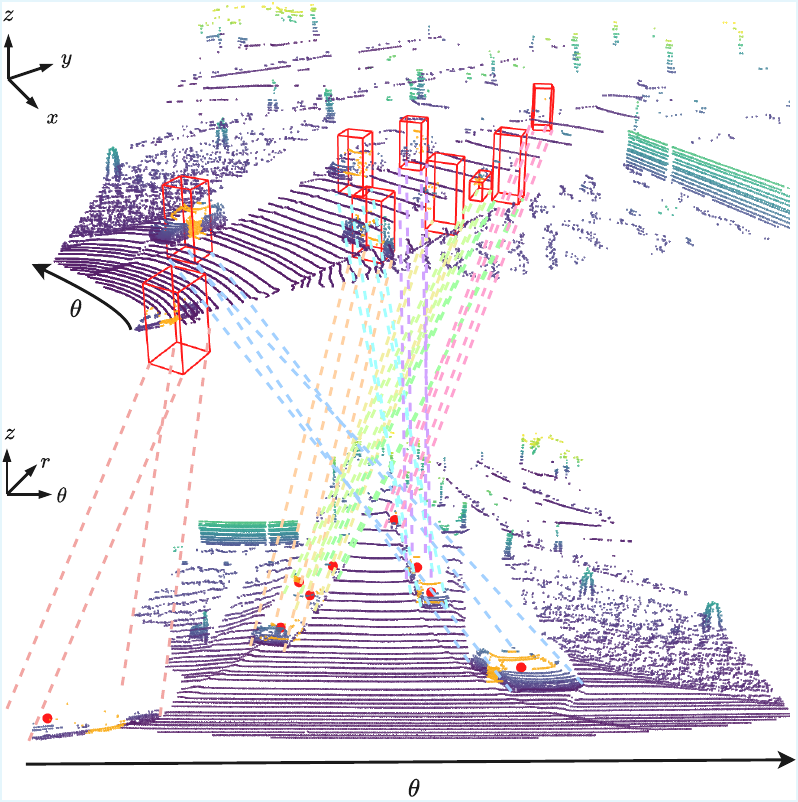}
    \caption{\textbf{Egocentric distortion in polar coordinates.} Illustration of object deformation. Objects of similar size are distorted differently relative to the ego vehicle.}
    \label{fig:distortion}
\end{figure}

\paragraph{Per-plane distortion.} To quantify local geometric distortion between Cartesian and cylindrical coordinate representations of a 3D point cloud (visualized in Fig. \ref{fig:distortion}), we compute a \textit{local area distortion} score based on pairwise distances in 2D projections. Given two sets of corresponding points \( \mathbf{P}_{cart}, \mathbf{P}_{cyl} \in \mathbb{R}^{M \times 2} \) (sampled from the full set if necessary), we define distortion as the standard deviation of the ratio of pairwise distances:

\[
\text{distortion} = \operatorname{std}_{m,n} \left( \frac{\|\mathbf{p}^{(2)}_m - \mathbf{p}^{(2)}_n\|_2}{\|\mathbf{p}^{(1)}_m - \mathbf{p}^{(1)}_n\|_2} \right), \quad m, n \in \{1, \dots, M\}
\]

where \( \mathbf{p}^{(1)}_m \in \mathbf{P}_{cart} \), \( \mathbf{p}^{(2)}_m \in \mathbf{P}_{cyl} \).

We evaluate this distortion over three 2D subspaces derived from the Cartesian and cylindrical representations of the point cloud:
\begin{itemize}
    \item In the \( (r, \theta) \) plane:  
    \[
\mathbf{p}^{(1)} = (x, y), \quad \mathbf{p}^{(2)} = (r, \theta)
\]
\[
r = \sqrt{x^2 + y^2}, \quad \theta = \arctan2(y, x)
\]

    \item In the \( (r, z) \) plane:
    \[
    \mathbf{p}^{(1)} = (x, z), \quad \mathbf{p}^{(2)} = (r, z)
    \]
    
    \item In the \( (\theta, z) \) plane:
    \[
    \mathbf{p}^{(1)} = (y, z), \quad \mathbf{p}^{(2)} = (\theta, z)
    \]
\end{itemize}

Each distortion score is computed using 1000 randomly sampled points to reduce computational cost. Finally, we normalize the distortions into percentages to understand the relative contribution of each subspace to the overall distortion:

\[
\text{percent}_{s} = \frac{\text{distortion}_{s}}{\sum_{t} \text{distortion}_{t}} \times 100\%, \quad s \in \{r\text{-}\theta, r\text{-}z, \theta\text{-}z\}
\]

\paragraph{Physical volume ERF.} One heuristic for measuring the compounding distortion with respect to network depth is how much physical volume represents ground truth foreground objects compared to background throughout the pipeline. As shown in Fig. ~\ref{fig:distortion} in the supplement, ground truth objects can be either stretched or compressed when in the egocentric polar view. As such, the proportion of foreground to background is also impacted by the polar coordinate system. Moreover, sparse convolutions exacerbate this distortion as stretched foreground regions—objects near the ego vehicle—reside in denser parts of the feature map and thus receive disproportionately high attention during feature aggregation. Conversely, compressed foreground regions—objects farther from the ego—occupy fewer voxels and risk being underrepresented. This imbalance in physical representation leads to a skewed effective receptive field (ERF), where certain spatial regions dominate the learned features, introducing compounding bias as the network deepens.

This spatial imbalance motivates the need for a more uniform treatment of physical space across depths. In contrast to polar representations, Cartesian coordinate systems preserve physical distances uniformly and exhibit a near-linear scaling between foreground ERF and total ERF as the network grows deeper. To mitigate the compounding distortion introduced by non-uniform spatial resolution in polar grids and sparse convolutions, we aim to design a method that more faithfully tracks the Cartesian ERF ratio—ensuring that foreground regions maintain consistent influence in the feature hierarchy while preserving the low-latency benefits of the polar view.

We measure the foreground effective receptive field ratio as the ratio of physical volume overlapping ground truth objects compared to overall physical volume of the sparse feature map. Specifically, let the voxel grid be defined in cylindrical coordinates with shape \( (N_\theta, N_r, N_z) \), and spatial bounds \([r_{\min}, r_{\max}]\), \([z_{\min}, z_{\max}]\), and \(\theta \in [-\pi, \pi]\). 

Let the voxel resolutions be defined as:
\[
\Delta r = \frac{r_{\max} - r_{\min}}{N_r}, \quad
\Delta z = \frac{z_{\max} - z_{\min}}{N_z}, \quad
\Delta \theta = \frac{2\pi}{N_\theta}.
\]

For a voxel at index \((i, j, k) \in \mathbb{Z}^3\), where \(r_j = r_{\min} + j \Delta r\), its physical volume is:
\[
V_{ijk} = \tfrac{1}{2} \left[ (r_j + \Delta r)^2 - r_j^2 \right] \Delta \theta \Delta z.
\]

Given a set of activated voxels \(\mathcal{A} \subseteq \mathbb{Z}^3\), define:
\[
V^{(L)}_{\text{total}} = \sum_{(i,j,k) \in \mathcal{A}} V_{ijk}, \quad
V^{(L)}_{\text{fg}} = \sum_{(i,j,k) \in \mathcal{A} \cap \mathcal{G}} V_{ijk},
\]
where \(\mathcal{G}\) denotes ground-truth foreground voxel indices.

Then the foreground ERF ratio is:
\[
R^{(L)} = \frac{V^{(L)}_{\text{fg}}}{V^{(L)}_{\text{total}}}.
\]

We evaluate the growth of the foreground ERF with respect to layer depth, where each layer is either a 3D downsampling convolution followed by a 3D upsampling convolution or two 2D decomposed convolutions followed by their corresponding upsampling convolutions. Between each convolution layer, we add batch normalization and ReLU activation, fully consistent with the pipeline within our PHiM block \ref{sec:phim}. Crucially, we find that despite poor scaling with deeper downsampling pipelines, decomposed convolutions closely track the Cartesian foreground ERF ratio within the range of 0 to 6 PHiM blocks. This motivates our choice for 6 stacked PHiM blocks in our final architecture.

\section{Additional results}

\paragraph{Efficiency metrics comparison with baseline Mamba.} To strengthen our efficiency claims from the main paper, we include a comparison of number of floating point operations per forward pass, number of parameters, and memory usage against the representative Voxel Mamba \cite{voxelmamba} in Tab. ~\ref{efficiency}. By decomposing 3D convolutions into separate 2D convolutions and removing positional encodings, PHiM saves almost 2x number of parameters. In conjunction, removing serialization curves in favor of streaming-native azimuthal ordering results in almost 2x reduction in peak memory usage. We report full-scan results on the Waymo Open dataset. 

\paragraph{Prediction refinement ablation.} We evaluate PFCF's performance on global scene predictions with varying number of buffered sectors in Fig. ~\ref{fig:refinement_line}. Because PFCF makes global predictions for each new input sector, it is able to gradually refine its predictions -- even on past spatiotemporal regions -- and approach full-scan performance while maintaining the streaming benefit. After the initial stage of "loading" the sector buffer, PFCF will continue performing at full-scan level. This is in contrast with previous streaming methods \cite{han, strobe, polarstream, partner} which only makes predictions on regions corresponding to the given input. As such, previous streaming methods always demonstrate constant global performance. 

\paragraph{Prediction refinement qualitative visualization.} Fig. ~\ref{fig:refine_qualitative} shows a qualitative visualization of the feature maps and output predictions for PFCF. Notably, while some ground truth objects are not detected initially (bottom row in the first and third frames), our sector buffer concatenates features together and captures the missing objects with the subsequent input sector (second and fourth frames).

\paragraph{Measuring efficiency.} Our efficiency comparison in Tab. ~\ref{efficiency} includes measurements on total number of floating point operations per forward pass, number of total parameters, parameter memory and peak memory usage. We use the ptflops \cite{ptflops} tool to measure floating point operations and built-in CUDA memory tracking in Pytorch to measure peak CUDA memory usage. We report averages over 1000 samples.

\section{Broader impacts discussion}
This work presents Polar-Fast-Cartesian-Full, a memory-efficient and low-latency architecture for LiDAR-based perception in autonomous systems. By reducing geometric distortion in egocentric coordinates and enabling streaming inference, our method may contribute to safer, more responsive autonomous driving technologies. The reduced computational footprint of PFCF also opens the door for deployment on lower-power edge devices, potentially democratizing access to advanced perception systems.

That said, improvements in 3D perception can also amplify capabilities in domains with dual-use concerns, such as surveillance or military applications. Moreover, safety and performance gains in perception alone do not guarantee equitable outcomes; system-level considerations -- such as dataset bias and deployment context -- remain critical. We encourage developers to evaluate downstream uses of our method with attention to fairness, transparency, and societal impact.

\end{document}